\documentclass[10pt,twocolumn,letterpaper]{article}

\usepackage{cvpr}
\usepackage{times}
\usepackage{epsfig}
\usepackage{graphicx}
\usepackage{amsmath}
\usepackage{amssymb}
\usepackage{multirow}
\usepackage{siunitx}
\newcommand{\tabincell}[2]{\begin{tabular}{@{}#1@{}}#2\end{tabular}}

\usepackage[breaklinks=true,bookmarks=false]{hyperref}

\cvprfinalcopy 

\def\cvprPaperID{****} 

\begin{document}

\title{PV-RCNN: The Top-Performing LiDAR-only Solutions for 3D Detection / 3D Tracking / Domain Adaptation of Waymo Open Dataset Challenges}

\author{\stepcounter{footnote}
	Shaoshuai Shi$^{1}$
	\quad Chaoxu Guo$^{1}$ 
	\quad Jihan Yang$^{2}$ 
	\quad Hongsheng Li$^{1}$	\vspace{0.2cm}\\
	$^1$Multimedia Laboratory, The Chinese University of Hong Kong \\
	$^2$The University of Hong Kong\\
	{\tt\small shaoshuaics@gmail.com \quad gus\_guo@outlook.com 
	\quad 
	jihanyang13@gmail.com \quad  hsli@ee.cuhk.edu.hk}
}

\maketitle

\begin{abstract}
	In this technical report, we present the top-performing LiDAR-only solutions for 3D detection, 3D tracking and domain adaptation three tracks in Waymo Open Dataset Challenges 2020. 
	Our solutions for the competition are built upon our recent proposed PV-RCNN 3D object detection framework. Several variants of our PV-RCNN are explored, including temporal information incorporation, dynamic voxelization, adaptive training sample selection, classification with RoI features, etc. A simple model ensemble strategy with non-maximum-suppression and box voting is adopted to generate the final results. By using only LiDAR point cloud data, our models finally achieve the 1st place among all LiDAR-only methods, and the 2nd place among all multi-modal methods, on the 3D Detection, 3D Tracking and Domain Adaptation three tracks of Waymo Open Dataset Challenges. Our solutions will be available at \url{https://github.com/open-mmlab/OpenPCDet}. 
\end{abstract}

\section{Introduction}
The Waymo Open Dataset Challenges at CVPR'20 are the highly competitive competition with the largest LiDAR point cloud dataset for autonomous driving. We mainly focus on the 3D detection track, which requires to localize and classify the surrounding objects of ego-vehicle in the 3D LiDAR point cloud scenes. With our proposed powerful 3D detector, we not only achieve 1st place on the 3D detection track among all LiDAR-only methods \cite{waymo_3ddet}, but also achieve the top performance on both the 3D tracking track and the domain adaptation track \cite{waymo_3dtrack, waymo_da}. 

\section{PV-RCNN: Solution to 3D Detection from Point Cloud}
3D detection with LiDAR point cloud is challenging due to its sparsity and irregular format. Previous methods generally either transform the point cloud to regular voxels for processing with regular convolution \cite{zhou2018voxelnet,yan2018second, shi2019part}, or directly estimate 3D bounding boxes with PointNet \cite{qi2017frustum,qi2017pointnet++} from raw point cloud \cite{qi2017frustum,shi2019pointrcnn}. Actually both voxel-based and point-based strategies have their advantages, where voxel-based strategy is generally more efficient and effective while point-based strategy has flexible receptive field and remains accurate point locations. 

Hence, we propose the PV-RCNN 3D detection framework to deeply integrate the voxel-based sparse convolution \cite{SubmanifoldSparseConvNet} and point-based set abstraction \cite{qi2017pointnet++} to bring the best from both of them. Our solutions for those competitions of Waymo Challenges are mostly built upon our PV-RCNN 3D detection framework.

\begin{figure*}
	\vspace{-5mm}
	\begin{center}
		\includegraphics[width=1.0\linewidth,height=6.2cm]{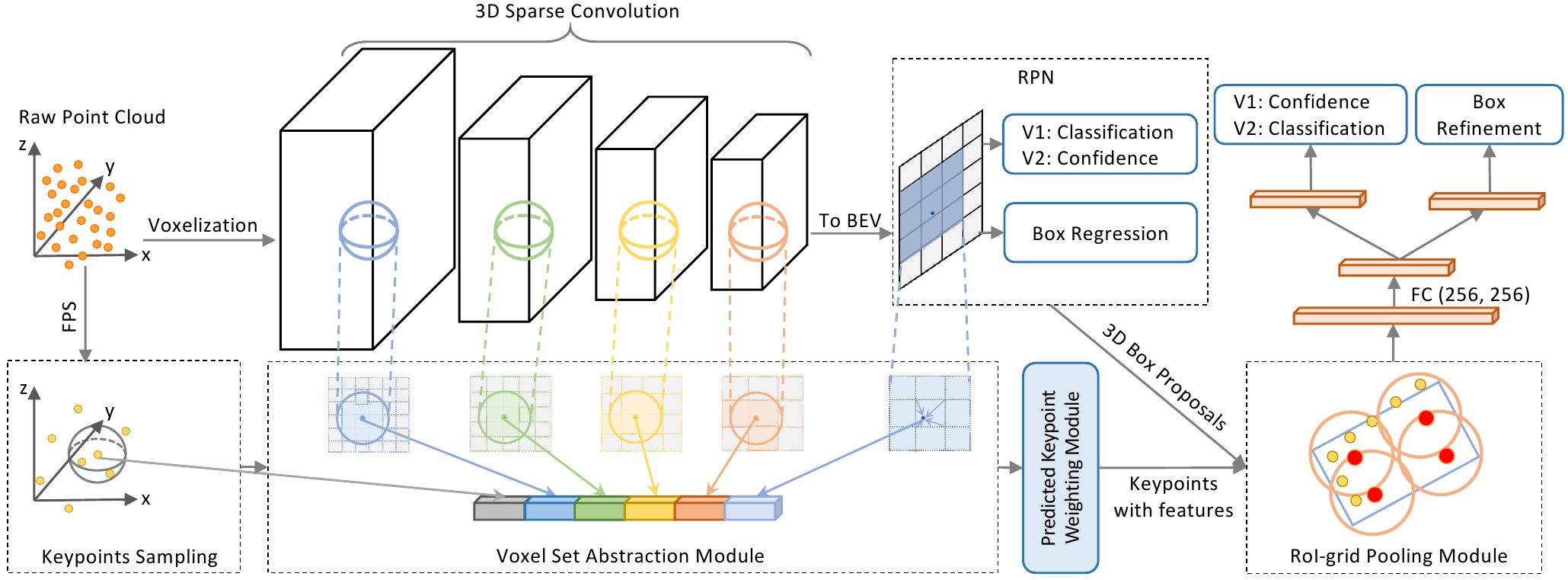}
	\end{center}
	\vspace{-0.2cm}
	\caption{The illustration of our proposed two variants of PV-RCNN 3D detection framework.}
	\label{fig:framework_total}
	\vspace{-0.2cm}
\end{figure*}

\subsection{Review of PV-RCNN 3D detection framework}
We first briefly review our PV-RCNN 3D detection framework proposed in our CVPR'20 paper \cite{shi2020pv}. The whole framework is illustrated in Fig.~\ref{fig:framework_total}, where the framework mainly has two stages, the voxel-to-keypoint scene encoding and the keypoint-to-grid RoI feature abstraction.

In the first voxel-to-keypoint scene encoding stage, we propose the Voxel Set Abstraction (VSA) layer to aggregate the multi-scale voxel features to a small set of keypoint features, where the keypoints are set as ball centers to aggregate the surrounding sparse voxel-wise features from multiple scales of 3D sparse-convolution-based backbone network. 
Hence, the keypoint features integrates the features from both the voxel-based sparse convolution and the point-based set abstraction, and also remain accurate point locations, which are especially important for the following fine-grained proposal refinement. 
In this stage, the high quality 3D proposals are also generated based on the predefined anchors on bird-view feature maps from the backbone.

Since the foreground keypoints should contribute more while background keypoints should contribute less in the following proposal refinement stage, we propose the 
Predicted Keypoint Weighting (PKW) module to further re-weight the keypoint features with extra supervision from point cloud segmentation.

In the keypoint-to-grid RoI feature abstraction stage, we propose the RoI-grid pooling module to aggregate the keypoint features to RoI-grid points. In contrast to the previous stage, here the RoI-grid points are set as the ball centers to group the features from surrounding keypoint features. Compared with previous 3D RoI pooling strategies \cite{shi2019pointrcnn,shi2020pv}, the proposed RoI-grid pooling scheme has larger receptive field and could even group the features of the surrounding foreground keypoints which are outside the 3D bounding box proposals to help to refine the 3D proposals. 

For more details of PV-RCNN 3D detection framework, please refer to our CVPR'20 paper \cite{shi2020pv}.

\subsection{Variants of PV-RCNN}
To further improve the 3D detection performance of PV-RCNN on the Waymo Open Dataset \cite{sun2019scalability}, we explored several modifications based on PV-RCNN framework for the following model ensemble. 

\noindent 
{\bf Incorporate last frame to get denser point cloud. }
The waymo dataset is composed of many temporal sequences where most frames could find the previous consecutive frames to compensate the information of current frame. 
We adopt a simple strategy to incorporate the previous frame to get a denser point cloud as the input of our detection framework. Specifically, 
denote the points of frame  at time step $t$ as 
$P^{t}=((x_1^t, y_1^t, z_1^t), (x_2^t, y_2^t, z_2^t), \cdots, (x_{n_t}^t, y_{n_t}^t, z_{n_t}^t))$ and the current frame is at time step $t$. We combine the points of frame $t$ and frame $t-1$ to get the final input points $\tilde{P^{t}}$ as follows: 
\begin{align}
	\tilde{P^{t}}=&\left((x_1^t, y_1^t, z_1^t, 0),  \cdots, (x_{n_t}^t, y_{n_t}^t, z_{n_t}^t, 0), \right. \\
	&\left. 
	(x_1^{t-1}, y_1^{t-1}, z_1^{t-1}, \delta), \cdots, (x_{n_t}^{t-1}, y_{n_t}^{t-1}, z_{n_t}^{t-1}, \delta) \nonumber 
	\right)
\end{align}
where $\delta$ is the time difference between frame $t-1$ and frame $t$ to discriminate these two frames ($\delta=0.1$ in waymo competition). This simple strategy are especially beneficial for the detection of small objects like pedestrian and cyclist.

\noindent 
{\bf Testing with dynamic voxelization.}
Dynamic voxelization is proposed in \cite{zhou2019end} to avoid information loss during voxelization process. In this competition, we only adopt the dynamic voxelization in the inference process while keeping the training process unchanged, which slightly improved the final detection accuracies (see Table~\ref{tab:waymo3d}).

\noindent 
{\bf Adaptive training sampling selection.}
Our previous PV-RCNN adopts anchor-based strategy for 3D proposal generation, where we need define separate anchors and hyper-parameters (\ie, IoU threshold of positive / negative samples) for each class.  Inspired by \cite{zhang2019bridging}, we adopt the similar adaptive training sampling selection strategy on our PV-RCNN framework to adaptively define the IoU threshold for each ground-truth object, which effectively removes most of hyper-parameters in anchor assignment.

\noindent 
{\bf Classification with RoI-aligned features.}
Due to the class-variant anchor definition, the object classification is conducted in the first proposal generation stage in our previous PV-RCNN, and the second stage only estimate the confidence of each 3D proposal. With the help of the above class-agnostic anchor definition, we propose another variant of our PV-RCNN framework to conduct the classification in the second stage with the RoI-aligned features by RoI-grid pooling (see Fig.~\ref{fig:framework_total}). It results in more accurate classification accuracy, and the experiments show that this variant of PV-RCNN could achieve higher detection accuracy on the pedestrian and cyclist categories (see Table~\ref{tab:waymo3d}).

\subsection{Model ensemble}
After obtaining multiple different 3D object detectors, we use an ensemble of these detectors for the final submission. In order to preserve all the possible predictions for model ensemble, non-maximum suppression is not applied on individual detector but instead the detection results of all detectors are merged and non-maximum suppression is applied once, which produces the NMS boxes. Then, we utilize the boxes before non-maximum suppression, \ie original boxes, to refine locations and dimensions of the NMS boxes into final boxes. The details are as follows:
\begin{equation}
b_{k}^{final, i} = \frac{1}{N} \sum_{j\in S}b_{k}^{original, j},  \\
k\in \{x,y,z,w,l,h\}
\end{equation}
where $S$ contains original boxes whose IoUs with NMS box $b^{nms, i}$ are higher than a thresh , $N$ is the number of the selected original boxes in $S$ and $k\in \{x,y,z,w,l,h\}$. The rotation and score of the original boxes are kept unchanged. We named this technique as 3D box voting. Finally the NMS box is replaced with the corresponding final box as the submission results.

Although the detection performance of Vehicle and Cyclist is improved significantly by 3D box voting while improvement of Pedestrian is unsatisfactory. Hence, we explore another model ensemble technique to improve the performance of Pedestrian. Instead of merging detection results of all detectors directly, we merge the detectors one by one. Specifically, we firstly select two detectors and the scores of detection boxes predicted by each detector are multiplied with a score weight, which is obtained by grid search on Waymo validation split to obtain the best Pedestrian performance. Then the detection boxes of two detectors are merged and non-maximum suppression is applied. The result after non-maximum suppression can be considered as result of a new detector and the procedure above is performed until the performance improvement of Pedestrian is minor. This technique is named as greedy ensemble. 

By employing 3D box voting and greedy ensemble to ensemble different detectors, the detection performance is improved significantly. The results are shown in Table \ref{tab:waymo3d}.

\section{Solution to 3D Tracking and Domain Adaptation of Point Cloud}
\subsection{Solution to 3D tracking challenge}
The goal of 3D object tracking is to find the correspondence between 3D boxes across frames given lidar and camera sequence. In this report we only focus on the lidar-only 3D object tracking. Considering the high performance of 3D object detection achieved by PV-RCNN, We use PV-RCNN as an off-the-shelf 3D object detector to obtain oriented 3D bounding boxes given the LiDAR
point cloud. In order to obtain object IDs of the 3D boxes, we borrow the idea from \cite{weng2019baseline}, where a combination of 3D Kalman filter and Hungarian algorithm
is used for state estimation and data association.
Although we utilize simple combination of off-the-shelf 3D object detector and tracker, it is extremely efficient and effective and our method rank 1st among all lidar-only methods and rank 2nd among multi-modal methods on the 3D object tracking leader board. The results are shown in Table \ref{tab:tracking}.

\subsection{Solution to domain adaptation challenge}
The domain adaptation challenge aims to adapt the 3D detector to the new location and new weather with limited labeled data. In this competition, to tackle this challenge, we adopt a straightforward strategy by directly fine-tuning our well-trained 3D detector on a small set of labeled data of target domain. Thanks to our strong 3D detector from source domain, as shown in Table~\ref{tab:da}, this simple strategy already achieves great performance on the target domain. Note that for the detection of cyclist on target domain, we directly adopt the 3D detector trained on source domain since the target domain has quite small number of cyclist.

\section{Experiments}

\noindent
{\bf Waymo Open Dataset} is the largest dataset with LiDAR point cloud for autonomous driving. 
For the 3D detection and 3D tracking tasks, there are totally 798 training sequences (around 160k samples),  202 validation sequences (around 40k samples) and 150 testing sequences (around 30k samples). Annotations are provided for only the training and validation set. 
For the domain adaptation task, there are totally 80 labeled training sequences (around 16k samples) and 20 labeled validation sequences (around 4k samples).  
In this competition, all of our models are only trained or fine-tuned on the training sequences. 

\noindent 
{\bf Training details.}
3D detection models are the most important parts in this competition, where we trained all models for 80 epochs from scratch with ADAM optimizer and learning rate 0.01. The cosine annealing learning rate strategy was adopted for decaying the learning rate. The models were trained with batch size 64 and 32 GTX 1080 Ti GPUs on the training set. For training with the proposal refinement network, we sample 128 proposals with 1:1 ratio for positive and negative proposals. The detection point cloud range is set to $x\in[-75.2, 75.2]$m, $y\in[-75.2, 75.2]$m and $z\in[-2, 4]$m, while the voxel size is $0.1\times 0.1 \times0.15$m.

We adopt the commonly used data augmentation for 3D object detection, including randomly flipping along $x$ and $y$ axes, global scaling with scaling factor sampled from [0.95, 1.05], randomly global rotation along $z$ axis with angle sampled from [$-\frac{\pi}{4}, \frac{\pi}{4}$], and the ground-truth sampling augmentation as in \cite{yan2018second}.

\begin{table*}
	\small 
	\vspace{-1mm}
	\begin{center}
		\scalebox{0.99}{
			\setlength\tabcolsep{2pt}
			\begin{tabular}{l|c|c|cc|cc|cc}
				\hline \multirow{2}{*}{~~~~~~~~~~~~~~~~~~~Setting} & 
				\multirow{2}{*}{\tabincell{c}{Eval\\Set}} &
				\multirow{2}{*}{\tabincell{c}{Training \\ Set}} & \multicolumn{2}{c|}{Vehicle} & \multicolumn{2}{c|}{Pedestrian}  & \multicolumn{2}{c}{Cyclist}\\
				& & & AP/APH (L1) & AP/APH (L2) & AP/APH (L1) & AP/APH (L2) & AP/APH (L1) & AP/APH (L2)\\
				\hline
				Baseline (original PV-RCNN) & val & $\sim$80k & 74.43/73.84	& 65.35/64.84	& 61.40/53.43	& 53.90/46.72 &	64.73/63.48 & 62.03/60.83\\
				+ incorporate last frame & val & $\sim$80k  & 74.65/74.06 & 65.59/65.07 & 	64.13/59.51	& 55.12/51.09 & 60.86/59.85 & 59.14/58.15\\
				+ dynamic voxelization testing & val & $\sim$80k  & 75.20/74.63 &	66.17/65.66 & 64.72/60.40 &	55.75/51.95 & 63.87/62.85 &	61.00/60.02	 \\
				+ 50 epochs with full training data &val &  $\sim$160k & 75.89/75.37 & 66.98/66.51 & 75.54/71.18 &	67.66/63.52	& 68.02/67.01 & 65.22/64.26 \\
				+ soft-nms &val & $\sim$160k  &  77.46/76.91 &	68.71/68.21 &  77.87/73.26 & 68.71/64.48 & 69.81/68.75 & 67.53/66.50 \\ 
				\hline 
				Classification with RoI features & val & $\sim$160k &75.81/75.36 & 66.91/66.50 & 78.92/75.12 & 69.84/66.36	& 73.24/72.12 & 70.41/69.34 \\
				\hline 
				Model ensemble & val & $\sim$160k & 78.70/78.13 & 70.13/69.61 & 81.72/77.65 &	72.80/69.02 & 74.70/73.49& 72.06/70.89
				\\ 
				\hline
				\hline 
				Model ensemble &test & $\sim$160k & 81.06/80.57 & 73.69/73.23 & 80.31/76.28&	73.98/70.16 & 75.10/73.84	& 72.38/71.16\\ 
				\hline 
			\end{tabular}
		}
	\end{center}
	\caption{3D detection performance of variants of PV-RCNN on the validation set and the final results on the test set of Waymo Open Dataset. Note that the baseline PV-RCNN are trained with 30 epochs on half training data.}
	\label{tab:waymo3d}
	\vspace{-2mm}
\end{table*}

\subsection{Results for 3D detection challenge}
As mentioned before, we explored several variants of PV-RCNN to improve the 3D detection accuracy. The detailed 3D detection results on the validation set are shown in Table~\ref{tab:waymo3d}. We could see that the performance on the validation set improves constantly by combined with more new features. The final submission is based on the model ensemble of the models of all variants of PV-RCNN framework mentioned before, and the ensemble validation and test results are also shown in Table~\ref{tab:waymo3d}.

\subsection{Results for 3D tracking challenge}
The submission for 3D object tracking is based on the model ensemble results of 3D object detection.
To generate the object ID for each 3D detection box, a combination of 3D  Kalman filter and Hungarian algorithm is used for state estimation and data association. The results of 3D object tracking on validation and test set of Waymo Open Dataset are presented in Table\ref{tab:tracking}.

\subsection{Results for domain adaptation challenge}
For the domain adaptation challenge, we fine-tuned the models of 3D detection challenge on the labeled data of target domain for 20 epochs. The final submission results are based on the model ensemble of fine-tuned models, except for the cyclist which we directly tested with the source models. The final domain adaption results are shown in Table~\ref{tab:da}.

\begin{table}
	\small 
	\begin{center}
		\scalebox{0.999}{
			\setlength\tabcolsep{2pt}
			\begin{tabular}{c|cc|cc}
				\hline 
				\multirow{2}{*}{\tabincell{c}{Category}} & \multicolumn{2}{c|}{Val Set} & \multicolumn{2}{c}{Test Set}\\
				& MOTA & MOTP & MOTA & MOTP \\
				\hline 
				Vehicle &57.20/53.58 & 16.73/16.73 & 60.97/57.73  & 16.09/16.14  \\
				Pedestrian & 55.98/55.23 & 31.20/31.27 & 55.32/53.80 & 31.63/31.63 \\
				Cyclist & 56.91/56.78 & 26.75/26.75  &55.13/55.07 & 27.14/27.14\\
				\hline
			\end{tabular}
		}
	\end{center}
	\caption{Performance of 3D tracking challenge on the validation and test set of Waymo Open Dataset.}
	\label{tab:tracking}
	\vspace{-0.2cm}
\end{table}

\begin{table}
	\small 
	\begin{center}
		\scalebox{0.92}{
			\setlength\tabcolsep{2pt}
			\begin{tabular}{c|cc|cc}
				\hline 
				\multirow{2}{*}{\tabincell{c}{Category}} & \multicolumn{2}{c|}{Val Set} & \multicolumn{2}{c}{Test Set}\\
				& AP/APH (L1) & AP/APH (L2) & AP/APH (L1) & AP/APH (L2) \\
				\hline 
				Vehicle & 71.93/70.88 & 62.14/61.19 & 71.40/70.70 & 59.67/59.08\\
				Pedestrian & 55.09/51.80 & 38.81/36.49 & 58.40/55.36 & 48.27/45.74 \\
				Cyclist & - &  - & 28.80/27.98 & 28.31/27.50 \\ 
				\hline
			\end{tabular}
		}
	\end{center}
	\caption{Performance of domain adaptation challenge on validation and test set of Waymo Open Dataset.}
	\label{tab:da}
	\vspace{-0.2cm}
\end{table}

{\small
\bibliographystyle{ieee_fullname}
\bibliography{egbib}

\begin{thebibliography}{10}\itemsep=-1pt

\bibitem{waymo_3ddet}
{3D Detection Leaderboard of Waymo Open Dataset Challenges}.
\newblock \url{https://waymo.com/open/challenges/3d-detection/}, Accessed on
  2020-06-01.

\bibitem{waymo_3dtrack}
{3D Tracking Leaderboard of Waymo Open Dataset Challenges}.
\newblock \url{https://waymo.com/open/challenges/3d-tracking/}, Accessed on
  2020-06-01.

\bibitem{waymo_da}
{Domain Adaptation Leaderboard of Waymo Open Dataset Challenges}.
\newblock \url{https://waymo.com/open/challenges/3d-tracking/}, Accessed on
  2020-06-01.

\bibitem{SubmanifoldSparseConvNet}
Benjamin Graham and Laurens van~der Maaten.
\newblock Submanifold sparse convolutional networks.
\newblock {\em CoRR}, abs/1706.01307, 2017.

\bibitem{qi2017frustum}
Charles~R. Qi, Wei Liu, Chenxia Wu, Hao Su, and Leonidas~J. Guibas.
\newblock Frustum pointnets for 3d object detection from rgb-d data.
\newblock In {\em The IEEE Conference on Computer Vision and Pattern
  Recognition (CVPR)}, June 2018.

\bibitem{qi2017pointnet++}
Charles~Ruizhongtai Qi, Li Yi, Hao Su, and Leonidas~J Guibas.
\newblock Pointnet++: Deep hierarchical feature learning on point sets in a
  metric space.
\newblock In {\em Advances in Neural Information Processing Systems}, pages
  5099--5108, 2017.

\bibitem{shi2020pv}
Shaoshuai Shi, Chaoxu Guo, Li Jiang, Zhe Wang, Jianping Shi, Xiaogang Wang, and
  Hongsheng Li.
\newblock Pv-rcnn: Point-voxel feature set abstraction for 3d object detection.
\newblock In {\em Proceedings of the IEEE Conference on Computer Vision and
  Pattern Recognition}, 2020.

\bibitem{shi2019pointrcnn}
Shaoshuai Shi, Xiaogang Wang, and Hongsheng Li.
\newblock Pointrcnn: 3d object proposal generation and detection from point
  cloud.
\newblock In {\em Proceedings of the IEEE Conference on Computer Vision and
  Pattern Recognition}, pages 770--779, 2019.

\bibitem{shi2019part}
Shaoshuai Shi, Zhe Wang, Jianping Shi, Xiaogang Wang, and Hongsheng Li.
\newblock From points to parts: 3d object detection from point cloud with
  part-aware and part-aggregation network.
\newblock {\em IEEE Transactions on Pattern Analysis and Machine Intelligence},
  2020.

\bibitem{sun2019scalability}
Pei Sun, Henrik Kretzschmar, Xerxes Dotiwalla, Aurelien Chouard, Vijaysai
  Patnaik, Paul Tsui, James Guo, Yin Zhou, Yuning Chai, Benjamin Caine, et~al.
\newblock Scalability in perception for autonomous driving: Waymo open dataset.
\newblock {\em arXiv}, pages arXiv--1912, 2019.

\bibitem{weng2019baseline}
Xinshuo Weng and Kris Kitani.
\newblock A baseline for 3d multi-object tracking.
\newblock {\em arXiv preprint arXiv:1907.03961}, 2019.

\bibitem{yan2018second}
Yan Yan, Yuxing Mao, and Bo Li.
\newblock Second: Sparsely embedded convolutional detection.
\newblock {\em Sensors}, 18(10):3337, 2018.

\bibitem{zhang2019bridging}
Shifeng Zhang, Cheng Chi, Yongqiang Yao, Zhen Lei, and Stan~Z Li.
\newblock Bridging the gap between anchor-based and anchor-free detection via
  adaptive training sample selection.
\newblock {\em arXiv preprint arXiv:1912.02424}, 2019.

\bibitem{zhou2019end}
Yin Zhou, Pei Sun, Yu Zhang, Dragomir Anguelov, Jiyang Gao, Tom Ouyang, James
  Guo, Jiquan Ngiam, and Vijay Vasudevan.
\newblock End-to-end multi-view fusion for 3d object detection in lidar point
  clouds.
\newblock {\em arXiv preprint arXiv:1910.06528}, 2019.

\bibitem{zhou2018voxelnet}
Yin Zhou and Oncel Tuzel.
\newblock Voxelnet: End-to-end learning for point cloud based 3d object
  detection.
\newblock In {\em Proceedings of the IEEE Conference on Computer Vision and
  Pattern Recognition}, pages 4490--4499, 2018.

\end{thebibliography}
}

\end{document}